\documentclass[journal]{IEEEtran}

\IEEEoverridecommandlockouts
\usepackage{color, colortbl}
\usepackage{adjustbox}
\usepackage{array}
\usepackage{graphicx}
\usepackage{hyperref}
\usepackage{booktabs}
\usepackage{amsmath}
\usepackage{amssymb}
\newcommand{\norm}[1]{\left\lVert#1\right\rVert}
\usepackage{float}
\usepackage[dvipsnames]{xcolor}
\usepackage{caption}
\usepackage{subcaption}
\usepackage{algorithm,algorithmic}
\usepackage{amsmath,amssymb,amsfonts}
\usepackage{multirow}
\usepackage{enumitem}
\usepackage{empheq}
\usepackage[sorting=none,style=numeric-comp]{biblatex} %Imports biblatex package
\def\BibTeX{{\rm B\kern-.05em{\sc i\kern-.025em b}\kern-.08em
    T\kern-.1667em\lower.7ex\hbox{E}\kern-.125emX}}

\begin{document}

\title{A Heuristic Informative-Path-Planning Algorithm for Autonomous Mapping of Unknown Areas}

\author{
\IEEEauthorblockA{Mobolaji O. Orisatoki, Mahdi Amouzadi, Arash M. Dizqah}\\
\IEEEauthorblockA{\textit{Smart Vehicle Control Laboratory (SVeCLab), School of Engineering and Informatics, University of Sussex, Brighton, UK}}\\
\{mo365, a.m.dizqah\}@sussex.ac.uk
}

\markboth{Journal of \LaTeX\ Class Files,~Vol.~14, No.~8, August~2021}%
{Shell \MakeLowercase{\textit{et al.}}: A Sample Article Using IEEEtran.cls for IEEE Journals}

\maketitle

%\markboth{IEEE Transactions on Robotics, VOL. ?, NO. ?, AUGUST 2023}{}
\markboth{AUGUST 2023}{}

\begin{abstract}
Informative path planning algorithms are of paramount importance in applications like disaster management to efficiently gather information through a priori unknown environments. This is, however, a complex problem that involves finding a globally optimal path that gathers the maximum amount of information (e.g., the largest map with a minimum travelling distance) while using partial and uncertain local measurements. This paper addresses this problem by proposing a novel heuristic algorithm that continuously estimates the potential mapping gain for different sub-areas across the partially created map, and then uses these estimations to locally navigate the robot. Furthermore, this paper presents a novel algorithm to calculate a benchmark solution, where the map is a priori known to the planar, to evaluate the efficacy of the developed heuristic algorithm over different test scenarios. The findings indicate that the efficiency of the proposed algorithm, measured in terms of the mapped area per unit of travelling distance, ranges from 70\% to 80\% of the benchmark solution in various test scenarios. In essence, the algorithm demonstrates the capability to generate paths that come close to the globally optimal path provided by the benchmark solution.

\end{abstract}

\begin{IEEEkeywords}
Non-Convex Areas, Completely Unknown Areas, Coverage Path Planning, Voronoi partitioning, Estimation of Information 
\end{IEEEkeywords}

\section{Introduction}

Mapping an unknown area by a robot is increasingly important in an ever-growing number of industries operating in environments detrimental to humans health, for example, mining, contaminated areas, fire and forestry, just to name a few \cite{PATLE2019582, Thanou2013}. These industries need robots that can autonomously plan their motion to explore unknown dangerous areas. Moreover, in such precarious environments, robots must be able to complete the given information-gathering task in the shortest possible time or distance due to their limited storage capacity of energy. Such path planning within unknown and complex (e.g., non-convex) areas remains an open research challenge.

%*Why is it important to map an unknown area.
Path planning for mapping has been studied by researchers in the past, however, a majority of these studies focus on cases where the map is known or partially known and the problem is to identify a priori unknown obstacles within the map. \cite{Aggarwal2020} provides a survey on these studies. 

For example, the authors in \cite{PATLE2019582,9561550, karaman2011sampling} provided a solution to such a path planning problem where the robots meet a given set of way-points on the known map with a minimum travelling distance while following the rules of the environment (i.e., no collision with the obstacles and walls). The works in \cite{9561550} and \cite{karaman2011sampling} also propose a coverage path planning (CPP) algorithm to determine these way-points ensuring the robot visits all areas of the map without overlapping paths. A CPP-led robot must generate a path that maximises the coverage of an area while avoiding obstacles. \cite{Galceran2013} provides a survey on different CPP algorithms.

The work in \cite{9561550} also introduces a technique to optimally move a robot within a set of predetermined points starting from an initial location to the neighbouring closest way-point one after another so that the total coverage path is traversed, moving the robot in a boustrophedon pattern \cite{fevgas2022coverage}. Alternatively, the authors in \cite{Arkin1994} proposed a solver to the CPP problems by generalising the travelling salesperson problem (TSP) in which the salesperson wishes to find the shortest path that visits a list of given clients within neighbourhoods where they are willing to meet. This technique is explored further in \cite{julia2012comparison}.

Moreover, the authors in \cite{Muralidharan2019} proposed a solution to a different but still relevant path planning problem where the robot needs to find a path on a graph with a minimum expected cost until success. Each node of the graph is assumed to carry a probability of failure and the mission success is defined as reaching a terminal node with zero probability of failure. The work also covers the cases where there are no such terminal nodes and the objective is to find a non-revisiting path with the minimum expected cost that traverses all the nodes carrying a non-unity probability of failure.

The aforementioned methods are non-adaptive and do not update the covering partitions to recalculate the travelling paths of robots as new obstacles are identified. The authors in \cite{Dutta2020} and \cite{Popovic2020}, on the other hand, proposed an informative path planning (IPP) where the robots continuously re-partition the known area based on newly identified obstacles. By solving an IPP problem, a robot generates an efficient path based on maximising the information that can be gathered. However, these works still focus on the efficient mapping problem of the partially known environment where the boundary of the area and the initial and final points are known.

Alternatively, the authors in \cite{julian2014} proposed an information-theoretic measure to estimate the mapping gains of each perceived cell, i.e., the rate of expanding the map if the robot moves to each cell. They proposed an algorithm to calculate this mapping gain and showed that a path-planning strategy which maximises this mapping gain eventually moves the robot to unexplored areas. 

The authors in \cite{Charrow2015} refined the measure in \cite{julian2014} using the Cauchy-Schwarz quadratic information (CSQMI), instead of the odds ratio of events, which they showed is more computationally effective for 3D mapping. The authors also proposed a path-planning algorithm to find out the next action that maximises the CSQMI normalised by the execution time of the action.

The works in \cite{Selin2019} combine the already applied frontier exploration planning (FEP) and receding horizon next-best-view planning (RH-NBVP) algorithms to explore unknown environments. The works show that such a combination recovers the disadvantages of the algorithms which are the slowness of FEP and the high risk of RH-NBVP getting stuck in large areas.

Similar to \cite{julian2014, Charrow2015, Selin2019}, the authors in \cite{Schmid2020} also considered the problem where the map is fully unknown in advance and proposed a sampling-based approach to explore such environments. In particular, they proposed a RRT*-based online path planning algorithm which is shown more effective as compared to other recent algorithms to explore more complex environments.  

Unlike the other research, the works in \cite{julian2014, Charrow2015, Selin2019, Schmid2020} address the problem with no prior knowledge of the boundaries and obstacles. However, they do not investigate the optimality of the generated path in terms of travelling distance or time. Since each segment of the path is generated online using the partially available measurements, an analysis showing how the set of locally generated paths is close to the global optimum solution is of paramount importance. Moreover, they do not consider the error between the measured distance of a point and its real location which causes the mapping of points beyond walls. This error requires embedding an explicit obstacle-avoidance mechanism into the path-planning algorithm. 

In summary, there is extensive prior research on path planning for gathering information. Many of the proposed algorithms only deal with partially unknown maps where only the location of obstacles is not known. The papers dealing with fully unknown environments, on the other hand, do not investigate the optimality of their solutions. This paper addresses these gaps by the following novel contributions to the knowledge:
\begin{itemize}
    \item A heuristic informative path planning (HIPP) algorithm that uses narrow-beam range detectors to create a map of the unknown non-convex areas with the shortest travelling distance. The proposed algorithm novelly approximates the mapping gain of all the identified cells online to locally find the next move of the robot which maximises the mapping gain with the shortest possible travelling distance. 
    \item An algorithm to approximate a near-global-optimal solution to the posterior problem as a benchmark to evaluate the results of the developed HIPP algorithm. In the posterior problem, the map is known to the planner which finds the shortest path for re-mapping of the known map. This algorithm can be used to calculate a benchmark for other IPP algorithms for mapping unknown environments.
    \item An optimality investigation of the resulting HIPP's solutions, indicating a linear increase of the reward over the majority of the travelling time which matches the performance of the benchmark solution.
\end{itemize}

The remainder of this paper is organised as follows: Section II provides a description of the involved components and their models. Section III introduces the proposed HIPP algorithm including the problem formulation and design of a solver. Section IV presents an algorithm to generate a benchmark solution which is then used in Section V to evaluate the efficacy of the proposed HIPP algorithm. This section provides the simulation results and discussions which is followed by a conclusion in Section VI. 

\begin{figure}[t]
   \centering
   \includegraphics[keepaspectratio,width=7.0cm]{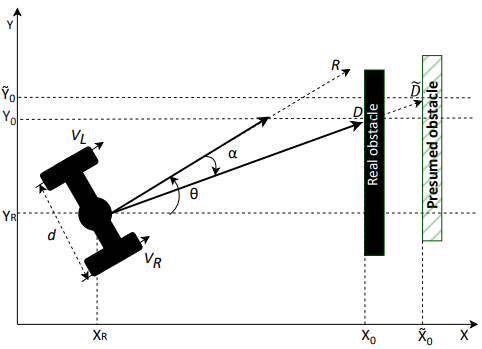}
   \caption{The single-axle robot in this research along with the measurement uncertainty of the LiDAR sensor which is used to develop the proposed HIPP. $D$ is the distance of the real obstacle but due to the sensor uncertainty, {$\tilde{D}$} is measured for the presumed obstacle. {$\alpha$} is the angle between the scanning ray and the x-axis of the robot and {$\theta$} is the heading angle of the robot. ${R}$ represents the nominal range of the sensor.}
  \label{UnknownScan}
  \vspace{-10pt} 
\end{figure} 

\section{System description }\label{systemdescript}
Fig.\ref{UnknownScan} shows a differential single-axle planar robot which is used throughout this paper to generate results. ${V_L}\,(m/s)$ and ${V_R}\,(m/s)$ are the linear speeds of the left and right wheels on the x-axis of the inertial frame. Any difference between these two speeds lays an inertial yaw rate on the vehicle, which is assumed to be the same as the variation of heading angle ${\theta}$ in the no-slip conditions.

Ignoring tyre slips, the kinematics of wheels are as follows:
\vspace{-10pt}
\begin{equation} 
 \begin{split}
   {V_L} = {\omega_Lr}; \;\;
   {V_R} = {\omega_Rr}.
 \end{split}
\end{equation}
where ${r}\,(m)$ is the radius of the wheels and ${\omega_i,\, i\in\{L, R\}}\,(rad/s)$ is the rotational speed of the left and right wheels.

The kinematics of the whole robot is then approximated as follows:
\vspace{-5pt}
\begin{equation}
 \label{robot_kinematics}
 \begin{split} 
   {V_c} = {\frac{V_R+V_L}{2}}; \;\;
   \dot\psi = {\frac{V_R-V_L}{d}}.
 \end{split}
\end{equation}
where $V_c\,(m/s)$ is the longitudinal speed of the robot in the x-axis of the moving frame of reference and $\psi\,(rad)$ is the yaw angle and hence $\dot{\psi}\,(rad/s)$ is yaw rate. As mentioned, it is assumed that $\dot{\psi} = \dot{\theta}$, ignoring the slips. 

It is accepted in this paper that the robot is able to estimate its global location $[X_r\,\, Y_r]^T$ within the map using a simultaneous localisation and mapping (SLAM) algorithm.

The robot is equipped with a 2D narrow-beam light detection and ranging (LiDAR) sensor which is located in its centre of gravity. This sensor emits 72 rays as per rotation and hence the angular scanning resolution is $\frac{360^{\circ}}{72}$ or 5$^{\circ}$. The reflected pulses are used to measure the distance of the obstacle, which is then summed up with the robot's location to estimate the obstacle's location. If there is no reflection, it assumes no obstacle within the nominal range ${R}$ of sensor \cite{baras2019autonomous}. 

This paper uses different models of the LiDAR sensor for the HIPP and posterior problems. While the sensor model must simulate an uncertain behaviour for the HIPP problem with respect to the range and direction of scans, the sensor model in the posterior problem, on the other hand, simulates laser scans with a certain range and direction. 

Fig.\ref{UnknownScan} shows an example measurement by an uncertain LiDAR sensor. While the location of an obstacle is measured as $[X_o\,\, Y_o]^T$, an additive white Gaussian noise (AWGN) is added to each measurement to imitate the uncertainty of the sensor:
\begin{equation}
 \begin{bmatrix}
  \tilde{X_o}\\
  \tilde{Y_o}
 \end{bmatrix} =
 \begin{bmatrix}
 X_o\\
 Y_o 
 \end{bmatrix} + \gamma;\;\; \gamma \sim \mathcal{N} (\begin{bmatrix} 0\\ 0  \end{bmatrix}, \begin{bmatrix} \sigma_{XX^2} & 0 \\ 0 & \sigma_{YY^2} \end{bmatrix}).
\end{equation}
where $\gamma$ is a multivariate random number with normal distribution. The uncertainty in the X and Y directions are independent with a variance of respectively $\sigma_{xx}$ and  $\sigma_{yy}$. The difference between these two values is a model depicting the uncertain deflection of the rays. 

If there is no obstacle, the reading $[X_o\,\,Y_o]^T$ will be $[R\cos(\theta)\,\,R\sin(\theta)]^T$ which is then summed up by an AWGN $\gamma$.

\begin{figure}[t]
   \centering
   \includegraphics[keepaspectratio,width=6.5cm]{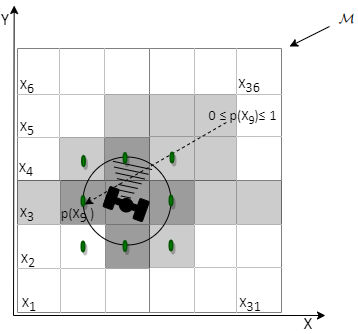}
   \caption{A 2-D occupancy grid map $\mathcal{M}$ with 36 cells and their numbering convention. $p(C_j) = 0$ means that cell $C_j$ is definitely empty. This figure also shows an example measurement of the ideal and uncertain sensors. A robot located in $C_{15}$ measures the darker grey cells when it uses an ideal sensor with the radius of $R=1 \times$ cell width and scans the lighter grey cells using the uncertain version of the sensor. In the ideal case, Only the cells whose centre points (indicated by green dots) within the circle are assumed to be scanned due to the quantization of the map.}
   \label{OGM_idealSensor}
  \vspace{-10pt} 
\end{figure}

Fig.\ref{OGM_idealSensor} compares sensor measurements for the ideal and uncertain cases. As mentioned, the sensor model for the posterior problem represents an ideal sensor with a certain range of $R$ (which equals $1\times$ cell width in Fig. \ref{OGM_idealSensor}). In this case, a cell is considered as scanned if the middle point of the cell,  noted as green dots, is within the range $R$ of the sensor.

The scanned environment is depicted as an occupancy grid map (OGM) \cite{collins2007occupancy}. Fig.\ref{OGM_idealSensor} illustrates an example map $\mathcal{M}$ consisting of $N=36$ cells $C_j,\, j\in\{1..N\}$ which are numbered in a boustrophedon pattern. 

Each cell in map $\mathcal{M}$ carries an integer $n_{ray}$ indicating the number of times the sensor rays pass through that cell. In other words, $n_{ray}$ increases every time a pulsed light passes through the cells in any direction. $n_{ray}$ is capped by $72$ since the cells where the robot is located and hence known as empty are ideally scanned 72 times when the sensor resolution is $5^\circ$. Using $n_{ray}$, a probability $P(C_j)$ is calculated for each cell as the likelihood of the cell not being empty, i.e., occupied by an obstacle. For the details, the reader is referred to Section \ref{HIPP}.

\section{The proposed HIPP algorithm} \label{HIPP}

This section presents the theoretical principles and design of the proposed HIPP algorithm that navigates a robot from an initial position to generate a 2D map of an unknown non-convex environment with the minimum possible travelling distance (i.e., consumed energy). It is important to note that the final destination of the robot is unspecified (as the environment is unknown) and exploration is completed when either a map with a given certainty is created or time reaches a threshold. 

Algorithm \ref{alg:cap1} summarises the proposed HIPP algorithm as a pseudo-code, including four major sections: i) generating an occupancy grid map in lines \ref{sensor_reading}-\ref{occupancygrid}; ii) estimating the information gain of each cell and finding the cells with the highest rewards in lines \ref{information}-\ref{maxentropy}; iii) finding the shortest paths from the current robot position passing through all the identified cells and select the first element of the path as the next destination, as in line \ref{shortestpath}-\ref{destination}(iv); and moving the robot along this path for one sampling time considering the potential obstacles in lines \ref{obstacle}-\ref{moverobot}. These sections are elaborated on in the following paragraphs. 

\vspace{-5pt}
\subsection{Map generation}
The uncertain reading of a LiDAR sensor at line \ref{sensor_reading} of Algorithm \ref{alg:cap1} is used to generate $\mathcal{M}_k$ containing the accumulative number of rays $n_{ray}$ passing through each cell up to sampling time $k$. The generated map is assumed to be a stochastic 2D occupancy grid map where the odds ratio (i.e., the robot's belief) for cell $C_j$ at each time $k$ is implicitly calculated as follows:
\begin{equation}
 \label{occupancy_prob}
   \begin{aligned}
       p(C_j; k) &= 1 - \frac{\sum_{l=1}^{k}\min(n_{ray}(C_j; l), 72)}{\min(\max(\mathcal{M}_k), 72) + \epsilon}; \: \forall j \in {1..N}.
   \end{aligned}
\end{equation}
where a small positive $\lim{\epsilon} \to 0$ is added to prevent the singularity in numerical calculations. Also, as explained in Section \ref{systemdescript}, using a sensor with a resolution of $5^\circ$, a cell $C_j$ is considered as a certainly empty cell once $72$ rays passed through it. Since the denominator of the robot's belief in (\ref{occupancy_prob}) is nonlinear, the iterative nature of (\ref{occupancy_prob}) is implicit and it is not trivial to decompose the odds ratio of the inverse sensor model.

\begin{figure}[t]
   \centering
   \includegraphics[keepaspectratio,width=7.0cm]{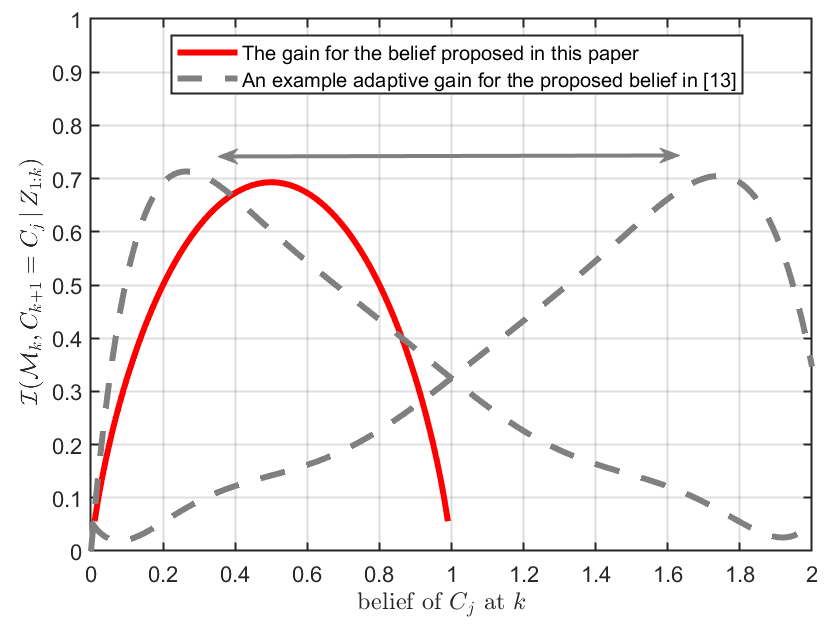}
   \caption{Mapping gain of a sample cell $C_j$ of map $\mathcal{M}_k$ based on the proposed formulation of belief in this paper (as red solid curve) as compared to the one in \cite{julian2014} (as dashed grey curves). The belief of cell $C_j$ is the odds ratio of the posterior of $C_j$ in \cite{julian2014} and is a number in $[0, \infty]$ where $0$ means that the cell is certainly empty and $\infty$ means that it is certainly occupied. The information gain \cite{julian2014} is dynamically adjusted based on the new reading guesses. This paper, on the other hand, suggests (\ref{occupancy_prob}) that gives a number in $[0, 1]$ where $1$ and $0$ mean that there is no information and the cell is certainly empty, respectively.}
   \label{InformationGain}
  \vspace{-10pt} 
\end{figure}

Apparently, $p(C_j)$ is a positive value between $[0,1]$ where $p(C_j) = 1$ means that there is no information about cell $C_j$ (or the cell is identified by all measurements to be occupied by an obstacle, which is unusual due to the uncertain nature of the sensor measurements). $p(C_j) = 0$, on the other hand, shows that cell $C_j$ is certainly an empty cell. If the truth is that the cell $C_j$ is empty, $p(C_j; k\to \infty)$ converges to $0$ because the numerator always grows faster than the denominator of (\ref{occupancy_prob}) till they are equal. If the truth is that the cell $C_j$ is occupied, on the contrary, the values of $p(C_j)$ will be close to $1$ and will not change except due to the measurement noise.

\algsetup{
  linenodelimiter = {    }
}

\begin{algorithm}[t]
\caption{A pseudo-code of the proposed HIPP algorithm}\label{alg:cap1}
\begin{algorithmic}[1]
    \STATE Initialise map $\mathcal{M}$ of arbitrary size $N$ with $n_{ray}(C_j) = -1$ and mutual information gain $\mathcal{I}$ of the map with $i(C_j) = 0$ (i.e., all cells are unknown). Initialise the initial location of robot $z_{orig}=[X_{r,0}, Y_{r,0}]$. Initialise sampling time $T_s$, mission time $T_f$, map reliability threshold $p_{th}$ and $k = 1$.\\[0.3cm]
    \WHILE {$k < T_f/T_s$ or $\exists \, j;\, p(C_j) \leq p_{th}$}
    \STATE $tmpMap$ $\gets$ depthScan($z_{orig}$) returns a $360^\circ$ sensor reading \label{sensor_reading} 
    \STATE $\mathcal{M} \gets$ element-wise $\mathcal{M}+tmpMap$ \label{accumulate_map}
    \STATE $p(C_j)$ $\gets$ updateProbability\Big($p(C_j)$\Big) $\forall j\in{1..N}$ using (\ref{occupancy_prob}) \label{occupancygrid}
    \STATE $\mathcal{I}$ $\gets$ updateInformationGain\Big($\mathcal{I}$, \{$p(X_0)..p(X_N)$\}\Big) penalising travelling distance too, as in (\ref{calc_information})\label{information}
    \STATE $cells$ $\gets$ highestInformation\Big($\mathcal{I}$\Big) using an $\epsilon$-greedy algorithm to select a predefined number of cells with the highest surrounding information\label{maxentropy}
    \STATE $\mathcal{P}_k$ $\gets$ shortestPath\Big($z_{orig}$, $cells$\Big)\label{shortestpath}
    \STATE $z_{dest} \gets \mathcal{P}_k(1)$ \label{destination}
    \IF {isObstacle\Big($z_{orig}$, $z_{dest}$\Big)} \label{obstacle}
        \STATE $z_{dest}$ $\gets$ RRT*\Big($z_{orig}$, $z_{dest}$\Big)
    \ENDIF
    \STATE $z_{orig}$ $\gets$ moveRobot\Big($z_{orig}$, $z_{dest}$\Big)\label{moverobot}
    \STATE $k$ $\gets$ $k + 1$\\[0.2cm]
    
    \ENDWHILE    
    \STATE returns $\mathcal{M}$ and $p(\mathcal{M})$
 \end{algorithmic}
\end{algorithm}
%-------geodesic--------------

\vspace{-5pt}
\subsection{Approximation of information}
The algorithm needs to estimate a mapping gain per each already mapped cell, which represents the potential expansion of the map if the robot moves to that cell. Having the previous travelled path $\mathbf{Z}_{1:k}$, this mapping gain at any time $k$ is equivalent to the mutual information $\mathcal{I}(\mathcal{M}_k; {C}_{k+1}\, |\, \mathbf{Z}_{1:k})$ between the map $\mathcal{M}_k$ and the next destination of robot ${{C}}_{k+1}$ from the current location of robot $z_{k}=[X_r\, Y_r]^T(k)$, which is defined as follows \cite{julian2014}:
\begin{equation}
 \label{calc_information_13}
   \begin{aligned}
       \mathcal{I}(\mathcal{M}_k; {{C}}_{k+1}\, |\, \mathbf{Z}_{1:k}) =& 
       \mathcal{H}(\mathcal{M}_k\, |\, \mathcal{O}({\mathbf{Z}_{1:k},{{C}}_{k+1})}) - \\
       &\mathcal{H}(\mathcal{M}_k\, |\, \mathcal{O}(\mathbf{\mathbf{Z}}_{1:k}))
   \end{aligned}
\end{equation}
where $\mathcal{H}(\mathcal{M}_k\, |\, \mathcal{O}(\mathbf{Z}_{1:k}))$ is the conditional entropy of map $\mathcal{M}_k$ knowing the previous observation $\mathcal{O}$ over the path $\mathbf{Z}_{1:k}$, and $\mathcal{H}(\mathcal{M}_k\, |\, \mathcal{O}(\mathbf{Z}_{1:k}:{{C}}_{k+1}))$ is a similar entropy when the next step destination at time $k$ (i.e., ${{C}}_{k+1}$) is also known. 

The authors in \cite{julian2014} formulated the belief that a cell is occupied (i.e., odds ratio) as a number between zero and infinity. They proposed an information gain function which is monotonically increasing before a peak where it switches to monotonically decreasing. It is shown in \cite{julian2014} that a path planning algorithm which maximises such an information gain eventually moves the robot to unexplored areas.

Moreover, the authors in \cite{julian2014} introduced a mechanism to adjust the information gain function by leaning its peak towards the low values of the belief or the other way around, as in Figure \ref{InformationGain}, depending on the guess of the next sensor reading for the cell. This helps to moderately correct an occupied belief by a new empty reading for a cell, and sharply strengthen an empty belief by a new reading which confirms the emptiness of the cell.

Conversely, this paper proposes (\ref{occupancy_prob}) to calculate belief as a positive number less than one. This helps to apply the Shannon formula to heuristically approximate the mutual information gain at each time $k$ between map $\mathcal{M}_k$ and moving the robot to any cell $C_j$ as follows:
\begin{equation}
  \label{calc_information1}
     \mathcal{I}(\mathcal{M}_k; {C}_{k+1}=C_j\, |\, \mathbf{Z}_{1:k}) \approx \mathcal{H}_s(C_j;k);\: \forall j \in \{1..N\}.
\end{equation}
where the Shannon entropy $\mathcal{H}_s$ is as follows:
\begin{equation}
  \label{Shannon_entropy}
  \begin{aligned}
       \mathcal{H}_s(C_j;k) = -\sum_{x\in\{0,1\}} p(C_j=x;k)\log p(C_j=x;k)
  \end{aligned}
\end{equation}

Figure \ref{InformationGain} illustrates the resulting curve (i.e., the solid red curve) of the proposed information gain with respect to the current belief of cell $C_j$. This formulation suggests that the cells which currently are the most uncertain (i.e., their belief is $0.5$) receive the maximum gain if the robot moves to them. 

To consider the potential gain of the surrounding environment of cell $C_j$, the proposed algorithm applies a $3\times3$ integrator kernel to magnify the high-information surroundings. 
The travelling distance is also taken into account as an additive penalty. The resulting information gain is then formulated as follows:
\begin{equation}
  \label{calc_information}
  \begin{aligned}
     \mathcal{I}_f(&\mathcal{M}_k; {C}_{k+1}=C_j\, |\, \mathbf{Z}_{1:k})  = \\
     &\Bigr[\mathcal{I}_f(\mathcal{M}_k; {C}_{k+1}=C_j\, |\, \mathbf{Z}_{1:k}) - \beta \norm{z_{k+1} - z_k}\Bigr] * \begin{bmatrix} 1 & 1 & 1 \\ 1 & 1 & 1 \\ 1 & 1 & 1 \end{bmatrix}
  \end{aligned}
\end{equation}
where $*$ represents the matrix convolution operator and $\beta \in [0,1]$ is the penalising factor of direct distance. $z_{k+1}$ is the location of centre of the cell $C_j$.
\vspace{-5pt}
\subsection{Path generation} 
This research proposes a path planning strategy that maximises $\mathcal{I}_f$ in (\ref{calc_information}). Whilst $I_f$ is an approximation of (\ref{calc_information_13}), it is already shown in \cite{julian2014} and section \ref{results} of this paper that such a path planning strategy eventually moves the robot to unexplored areas. At each sampling time $k$, having the information gain (\ref{calc_information}) for each cell of the map, The proposed strategy selects a predefined number of the cells with the highest gain. An $\epsilon$-greedy algorithm is used for this selection to add elements of exploration as well. A travelling salesperson problem with the return is, then, formulated and solved to find the shortest path $\mathcal{P}_k$ starting from the current location of the robot and passing through all the selected cells before returning back.

\subsubsection{Robot control and potential obstacle avoidance} The proposed HIPP algorithm uses (\ref{robot_kinematics}) to calculate velocities of the left and right wheels to move the robot from its current position to the first destination of the calculated path $\mathcal{P}_k$ (i.e., $\mathcal{P}_k(1)$). The velocities are applied to the robot for a sampling time $T_s$.

The proposed HIPP is equipped with a rapid exploring random tree star (RRT*) algorithm for obstacle avoidance. HIPP examines if there is an obstacle between the current location of the robot and the desired destination $\mathcal{P}_k(1)$. If an obstacle is detected, the RRT* is applied to find a path which avoids collision. Then, the first element of this path is considered the next destination of the robot. RRT* is an efficient sample-based algorithm that can search non-convex and high-dimensional environments by constructing spatial filling trees randomly during exploration \cite{8329210,9194248}.  

\section{The Proposed Algorithm to Calculate Benchmark Solution} \label{posteriorsection}

This section develops a novel benchmark solution to evaluate the performance of the proposed HIPP in section \ref{HIPP}. Principally, this benchmark is the optimal solution of the posterior mapping problem where the robot remaps a known environment using a limited-range ideal sensor as in Fig. \ref{OGM_idealSensor}.

The solution to the posterior problem is the shortest path to remap the area, in which the robot is given an initial position but no final destination. A formulation of the posterior problem is provided in (\ref{OCP}) as a minimum-time optimal control problem which is strongly non-convex. 

\begin{subequations} \label{OCP}
\begin{alignat}{10}
&\mathbf{U}^*(z_{init},G,R) = \text{arg}\:\underset{\substack{\mathbf{U}, N}}{\text{maximise}}\quad J(map,\mathbf{X},\mathbf{U}) := \notag \\
      &\quad \quad \quad \quad \sum_{k=1}^{N}\Big[\alpha f_I\big(\mathcal{M}_k, u_k \,|\, \mathbf{Z}_{1:k}\big)- \beta f_d\big(z_k,u_k\big)\Big] - \\ \notag
      &\quad \quad \quad \qquad  \mu N + \gamma\big|\mathcal{M}_N\big|\\
& \quad \quad \text{where:}\notag\\
& \quad \quad \quad f_I\big(\mathcal{M}_k; u_k \, | \,\mathbf{Z}_{1:k}\big) = \left|\mathcal{M}_k \cup sur\big(u_k,R\big)\right|-\left|\mathcal{M}_k\right|\\
& \quad \quad \quad f_d\big(z_k,u_k\big) = \norm{z_k- u_k} ^2\\
& \quad \quad \text{s.t.:}\notag\\
& \quad \quad \quad\quad \mathcal{M}_{k+1} = \mathcal{M}_k \cup sur\big(u_k, R\big); \:\: \mathcal{M}_1 = \emptyset\\
& \quad \quad \quad \quad z_{k+1} = u_k; \:\: z_1=z_{init}\\
& \quad \quad \quad \quad \mathbf{V}_{k+1} = \mathbf{V}_k \cup u_k\\
& \quad \quad \quad \quad u_k \in \mathbf{U}_k(G) \backslash \mathbf{V}_k
\end{alignat}
\end{subequations}
where $G(\mathbf{U}, \mathbf{E})$ is a graph of all feasible locations $\mathbf{U}$ of the robot on the map while every two locations with a line-of-sight are connected by an edge $e \in \mathbf{E}$ representing their Euclidean distance. $\mathbf{U}_k(G) = \{ u\mid u \in \mathbf{U}$; there is a direct link from $z_k$ to $u$ in the graph $G$ \}, $V_k$ =\{ $u\mid u \in U$; $u$ is an already visited vertex up to time $k$\}, $sur(u_k, R)$ is the set of the surrounding cells of $u_k$ with a sensor of range $R$. $\alpha$ and $\beta$ are arbitrary convex coefficients. $|.|$ returns the cardinality of a set.

The objective of the problem (\ref{OCP}) implies that the control strategy must find the shortest path that maximises the size of the generated map $\mathcal{M}$ in a minimum time $N$. It is worth noting that the final destination is not known which makes the problem even more complex. 

To solve the problem (\ref{OCP}), this paper proposes a heuristic solver that finds a near-global-optimal solution. Algorithm \ref{alg:cap2} summarises the proposed solver as a pseudo-code. As shown, the solver combines a CPP (Algorithm \ref{alg:cap2}, lines \ref{voronoiline}-\ref{CPPline}) and a TSP (Algorithm \ref{alg:cap2}, lines \ref{genGraph}-\ref{TSPline}) algorithm. The CPP algorithm optimally locates a given number of way-points on the map to maximise the coverage with a given sensor range $R$. The TSP algorithm, then, finds the shortest path from the initial point to an unknown destination that passes all the dispersed way-points. The details of these algorithms are provided in the remainder of this section. 

\begin{algorithm}[t]
\caption{A pseudo-code of the proposed heuristic solver to find the near-to-optimal solution of the posterior problem (\ref{OCP}), i.e, a benchmark solution. }\label{alg:cap2}
\begin{algorithmic}[1]
    \STATE Initialise the map $\mathcal{M}$, number of way-points $N_{wp}$, initial location of robot $z_{orig}=[X_{r,0}, Y_{r,0}]$, required number of generations $N_{gen}$ and k=1.\\[0.3cm]
    
    \STATE $\mathcal{S}_0$ $\gets$ randomSeedsOnMap($\mathcal{M}$, $N_{wp}$)
    \WHILE {$k \leq N_{gen}$}
        \STATE $\mathcal{V} \gets$ createVoronoiPartitions\big($\mathcal{M}, \mathcal{S}_{k-1}$\big) \label{voronoiline}
        \FOR{$i=1:N_{wp}$}
            \STATE $\mathcal{S}_k(i) \gets$ updateWayPoints\big($\mathcal{S}_{k-1}(i)$, $\mathcal{V}$\big) using the modified version of the algorithm in \cite{Thanou2013}, proposed in section \ref{CPP} \label{CPPline}
            \IF {isObstacle\Big($\mathcal{S}_{k}(i)$, $\mathcal{S}_{k-1}(i)$\Big)}    \label{obstacle}
                \STATE $\mathcal{S}_k(i)$ $\gets$ RRT*\Big($\mathcal{S}_{k-1}(i)$, $\mathcal{S}_{k}(i)$\Big)
            \ENDIF
            \STATE $\mathcal{S}_{k}(i)$ $\gets$ moveRobot\Big($\mathcal{S}_{k-1}(i)$, $\mathcal{S}_{k}(i)$\Big)
        \ENDFOR\\
        $k$ $\gets$ $k+1$
    \ENDWHILE
    \STATE $G(\mathcal{S}_{N_{gen}}, E) \gets$ generateGraph\big($\mathcal{S}_{N_{gen}}$\big) with the edges connecting way-points with line-of-sight by their Euclidean distance \label{genGraph}
    \IF {there is no connected graph}
        \STATE increase $N_{wp}$ and run the algorithm again
    \ENDIF
    \STATE $\mathcal{P} \gets$ shortestPath\big($z_{orig}$, $\mathcal{S}_{N_{gen}}$\big) calculates exhaustively the shortest path from $z_{orig}$ passes all the way-points as in section \ref{TSP}  \label{TSPline}
    \STATE returns $\mathcal{P}$
\end{algorithmic}
\end{algorithm}

\vspace{-5pt}
\subsection{The Proposed CPP Algorithm}\label{CPP}
\begin{figure}[t]
\centering
\begin{subfigure}[b]{0.227\textwidth}
        \includegraphics[keepaspectratio,width=4.2cm]{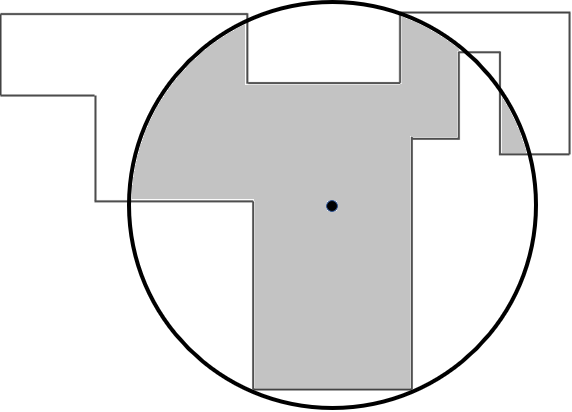}
         \caption{}
         \label{EuclideanDisc}
\end{subfigure}
\begin{subfigure}[b]{0.227\textwidth}
        \includegraphics[keepaspectratio,width=4.2cm]{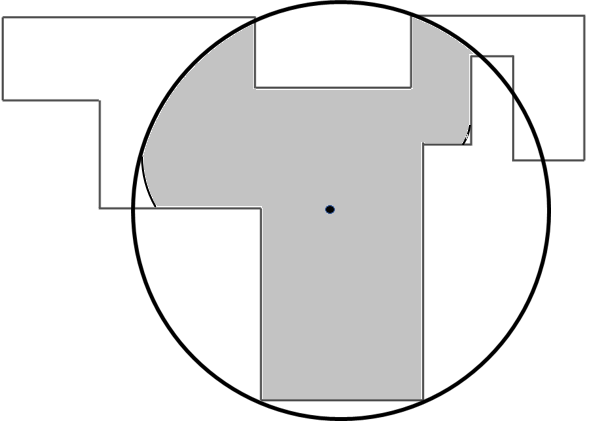}
        \caption{}
         \label{GedesicDisc}
\end{subfigure}\\
\begin{subfigure}[b]{0.427\textwidth}
        \centering        \includegraphics[keepaspectratio,width=4.0cm]{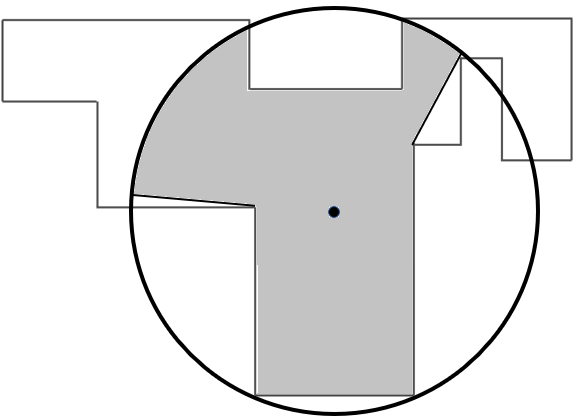}
         \caption{}
         \label{VisisbleDisk}
\end{subfigure}
\caption{ Different disc techniques can be used to represent the ideal sensing region within a non-convex map for calculating the benchmark solution. These techniques are a) Euclidean, b) range-limited visibility and c) geodesic. The black dots show the location of the sensor and the shaded areas depict the scanned areas. While the geodesic disc represents the area that is scanned by a non-line-of-sight radio sensor, the range-limited visibility technique is suitable for modelling the camera. As shown in (a), Euclidean discs are not suitable for non-convex areas because of including non-reachable points.}
\label{fig:discs}
\vspace{-5pt}
\end{figure}

This section proposes a modified version of the CPP algorithm in \cite{Thanou2013} and \cite{Stergiopoulos2015} to maximise the coverage of a non-convex map, such as in Fig. \ref{fig:discs}, with a finite number of robots equipped with sensors of range $R$.  

The CPP algorithm in \cite{Thanou2013} and \cite{Stergiopoulos2015} randomly seeds a finite number of moving robots across a map. Then, it creates Voronoi partitions \cite{Bakolas2015} for the seeds based on a geodesic proximity measure and hence decomposes the coverage problem among the finite number of robots. Unlike the original algorithm, this paper assumes that the sensors are of the line-of-sight type (e.g., LiDAR) and therefore the original geodesic technique is replaced by the range-limited visibility disc.

After partitioning the area, the CPP algorithm moves the robots using a free-arc technique iteratively to minimise the overlap between the sensing range of the robots. The details of these steps are explained in this section. 

\begin{figure}[t]
\centering
    \includegraphics[keepaspectratio,width=7.2cm]{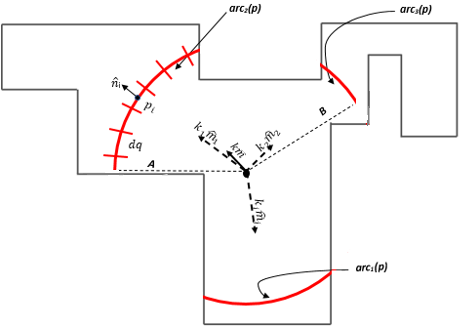}
     \caption{The proposed CPP algorithm, as part of Algorithm \ref{alg:cap2}, uses the free-arc method in \cite{Stergiopoulos2015} to calculate the desired movement $k\hat{m}$ of a robot within a non-convex map to maximize its sensing coverage. The sensor readings are modelled as a range-limited visibility disc (indicated by red arcs). $k\hat{m}$ is the sum of $k_j\hat{m_j}$ vectors which are calculated by applying (\ref{free_arc}) to each arc of the disc. Arcs are segmented with an equal length $dq$. $\hat{n_i}$ indicates a normal vector at the centre point $p_i$ of segment $i$.}

\label{fig:freeArc}
\vspace{-10pt}
\end{figure}

%----------------------------
 \begin{figure}[t]
 \centering
 \begin{subfigure}[b]{0.23\textwidth}
         \includegraphics[keepaspectratio,width=4.2cm]{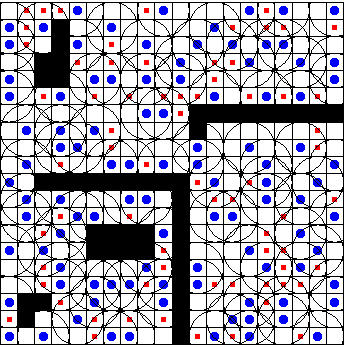}
         \caption{}
          % \caption{Coverage path planning (CPP)}
          \label{CPPMap}
 \end{subfigure}
 \begin{subfigure}[b]{0.23\textwidth}
         \includegraphics[keepaspectratio,width=4.2cm]{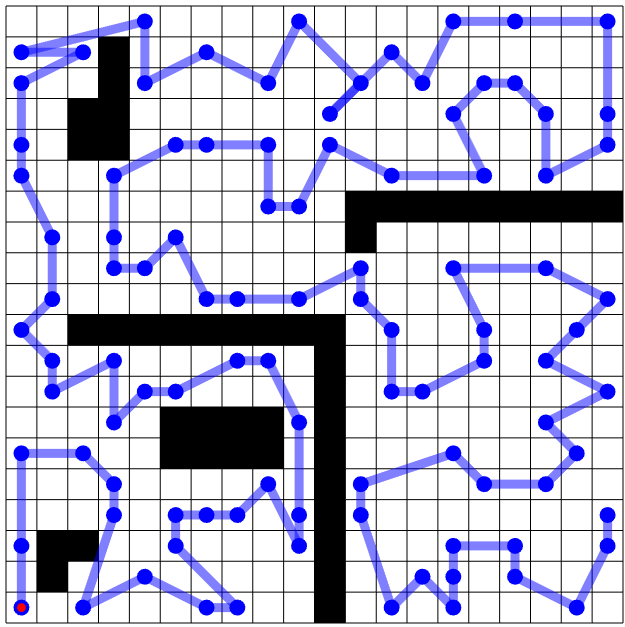}
          \caption{}
          % \caption{Traveling Salesman's Problem (TSP)}
          \label{TSPMap}
 \end{subfigure}
 \caption{The results generated by Algorithm \ref{alg:cap2}, i.e., a benchmark solution, for a sample known map, including (a) the final location of nodes (in blue), derived iteratively from the initial random positions (in red) by the proposed CPP algorithm to maximize the coverage. The black discs represent the coverage of the ideal measuring sensor of the nodes; and (b) the shortest path derived by the proposed TSP algorithm which starts from the initial location of the robot (indicated by a red circle) and passes through all the nodes without prior knowledge of the destination node.}
\label{fig:CPPTSP}
\vspace{-5pt}
\end{figure}

\subsubsection{Proximity measurements}%Part of CPP------
It is important to consider the non-convexity of the map during proximity measurement between nodes. Euclidean distance measure is not suitable for non-convex areas since it may include non-reachable parts of the map within a partition, as in Fig. \ref{EuclideanDisc}.

To address this issue, the authors in \cite{Stergiopoulos2015} used a geodesic disc technique, as in Fig. \ref{GedesicDisc}, where the proximity is measured by a non-direct but still reachable distance between points. This technique fits well to the radio sensors which are able to sense non-line-of-sight points.

As mentioned before, this paper uses a LiDAR sensor and hence employs a range-limited visibility disc technique as the proximity measure, where only the visible points within the sensor range $R$ belong to the partition \cite{ganguli2009multirobot} (Fig. \ref{VisisbleDisk}).

\begin{figure}[t]
\centering
\begin{subfigure}[b]{0.227\textwidth}
        \includegraphics[keepaspectratio,width=4.1cm]{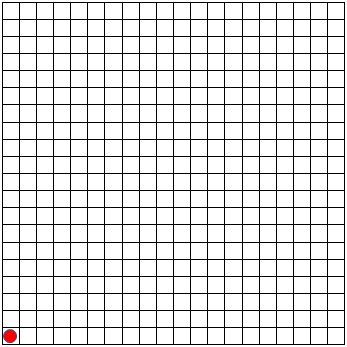}
         \caption{}
         \label{EmptyMap1}
\end{subfigure}
\begin{subfigure}[b]{0.227\textwidth}
         \includegraphics[keepaspectratio,width=4.1cm]{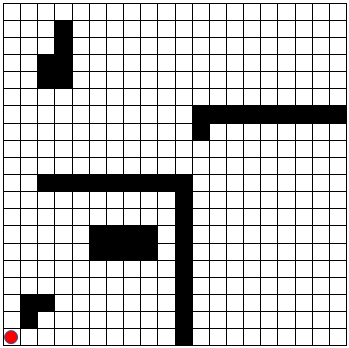}
         \caption{}
         \label{EmptyMap2}
\end{subfigure}\\
\begin{subfigure}[b]{0.327\textwidth}
        \includegraphics[keepaspectratio,width=4.3cm]{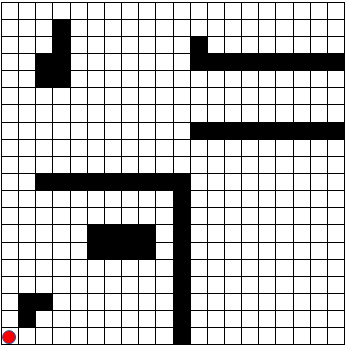}
        \centering
        \caption{}
        \label{EmptyMap3}
\end{subfigure}

\caption{Three different scenarios, including a $20 \times 20$ cells map and an initial location of the robot, is used in this study to test the performance of the proposed HIPP algorithm as compared to the benchmark solution; where (a) is Test Scenario 1 which is an open space without obstacles; (b) is Test Scenario 2 consists of five different obstacles and 355 available cells; and (c) is Test Scenario 3 which includes 346 free cells with an increased complexity by adding narrow passages. }
\label{fig:emptyMaps}
\vspace{-10pt}
\end{figure}

\subsubsection{Free arc method}%Part of CPP-----------
This paper uses the proposed free-arc method in \cite{Stergiopoulos2015} to calculate the vector of movement of the way-points in a non-convex environment. The method creates Voronoi partitions for the way-points. It draws a circle around the way-points with the radius of the sensor range $R$. Parts of the circle's edge outside of the Voronoi partition of the way-point are discarded and only the parts inside the partition are used to calculate the velocity (speed and its direction) of movement of the way-point.

Fig.\ref{fig:freeArc} illustrates the details of how the free-arc method calculates the velocity of a way-point for the case when the way-point has three eligible arcs (red arcs). As shown, the movement vector $k\hat{m}$ of each way-point is the vector sum of the resulting velocity $k\hat{m}_j$ of each arc $j$. The velocity due to each arc is calculated by splitting the arc into segments with an equal length of $dq$ and a centre point of $p_i$. Therefore, the velocity of the required movement of a way-point is calculated as follows:
\begin{equation}\label{free_arc}
k\hat{m}=\sum_{j=1}^{N_{arcs}} \int_{p_i \in arc_j} \phi\, \hat{n_i} dq
\end{equation}
where $N_{arcs}$ is the number of eligible arcs, $\hat{n}_i$ is the unit normal vector to the segment with the centre $p_i$, $dq$ is the length of segments and $\phi$ is a gain indicating the importance of each segment (that can be one for all segments). In other words, each segment $p_i$ affects the way-point velocity by a factor $\phi dq$ in the direction of $\hat{n_i}$.

Fig. \ref{CPPMap} shows the results of applying the free-arc algorithm to move the initially positioned red way-points to the better-positioned (i.e., smaller overlapped sensor ranges or equivalently larger covered area) blue points.

\vspace{-5pt}
\subsection{The Proposed TSP Algorithm}\label{TSP}
The optimally relocated way-points by the free-arc algorithm are used to create a graph connecting all the points with line-of-sight through their Euclidean distance. In other words, only the way-points which can be connected by a line without having an obstacle in between are connected in the resulting graph. 

The constructed graph is then used to exhaustively find the destination to which the path from the robot's original location passing through all the way-points is the shortest amongst the similar paths to all other destinations on the graph. A TSP is formulated and solved for each vertex of the graph \cite{5364611} and the shortest is chosen as a close-to-global solution of the original IPP problem.

Fig. \ref{TSPMap} illustrates an example optimum path generated by the proposed TSP algorithm knowing the initial location of a robot (red point) and using the relocated blue way-points in Fig. \ref{CPPMap}.

\section{Simulation Results and Discussions}
\label{results}
Three dissimilar maps in Fig.\ref{fig:emptyMaps} are used as test scenarios to evaluate the adaptation and efficacy of the proposed HIPP algorithm to different environments. Fig. \ref{EmptyMap1} is an empty map, whilst Fig. \ref{EmptyMap2} is a complex map with multiple rooms and obstacles. Fig. \ref{EmptyMap3}, as compared with Fig. \ref{EmptyMap2}, contains narrow dead-end passages. The results and performance of the proposed HIPP algorithm are evaluated and benchmarked against the solutions to the proposed posterior problem. 

Test scenarios are labelled as 1 to 3 which are aligned with Fig.\ref{EmptyMap1} to Fig.\ref{EmptyMap3}. Without loss of generality, the initial position of the robot in all scenarios is assumed identical, as the red circles in the left-bottom corner of Fig. \ref{fig:emptyMaps}.

The comparison measure is the number of identified cells on the OGM map normalised by the travelling distance. However, to make a realistic comparison, an equivalent number of identified cells is introduced in this section for the results of HIPP, which is the number of identified cells calculated by placing the ideal sensor of Fig.\ref{OGM_idealSensor} at the generated way-points by HIPP. Hence, a realistic comparison measure is the equivalent number of identified cells per unit of travelling distance.

Moreover, this section provides an optimality analysis where the accumulative performance of the proposed HIPP over the travelling time is benchmarked against the solution of the proposed posterior problem. In other words, it shows whether the number of identified cells increases linearly by the travelling time. 

The confidence of the mapped cells is imperative to the performance of the HIPP algorithm as the sensor's range and direction are uncertain. This confidence is measured by the number of times a cell is sensed. For the map to have high confidence, this paper only considers a cell empty if the probability of being sensed as empty is equal to or more than $20\%$. It means that a cell must be scanned at least 14 times as an empty cell to be considered so.

The HIPP and solver of the posterior problem run ten times for each test scenario. Table \ref{tab5} summarises the statistics of the resulting total travelling time (TTD), the number of mapped cells, the number of identified cells per unit of travelling distance(TD), the equivalent number of mapped cells, and the equivalent number of mapped cells per unit of travelling distance(TD). As explained before, for efficiency, the normalised equivalent number of identified cells (i.e., the latter column of the table) by Algorithm \ref{alg:cap1} is compared to the normalised number of identified cells (i.e., the third column of the table) by Algorithm \ref{alg:cap2}. 

\begin{figure}[!]
\centering
    \includegraphics[keepaspectratio,width=7.0cm]{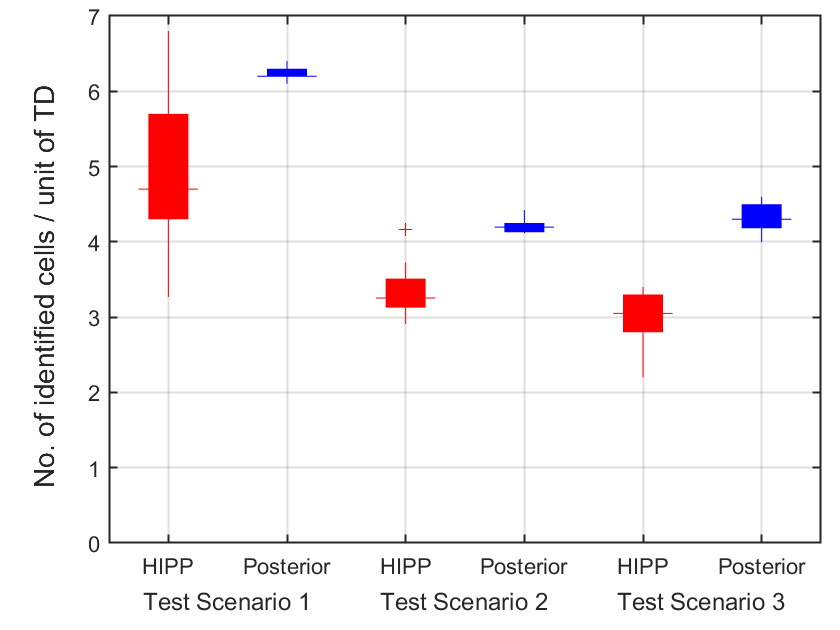}
     \caption{A graphical representation of the provided results in Table \ref{tab5} showing that in average HIPP, with no a priori knowledge of the map, effectively follows the benchmark solution, where the map in known in advance for different test scenarios in Figure \ref{fig:emptyMaps}.}
     \label{Results barchart}

\label{fig:barchart}
\end{figure}

Fig.\ref{fig:barchart} visualises this comparison as the box plots. As shown, the average performance of the proposed HIPP reaches $80\%$ and $70\%$ of the proposed benchmark for test scenarios 1 or 2 and the more complex test scenario 3, respectively. It also shows how the proposed solver finds a sub-optimal (if not close to global optimal) solution to the posterior problem. It is worth noting that in test scenario 1 the HIPP outperformed on one occasion which is due to an empty map and uncertain sensor. The results also show consistency in HIPP results, where mapping cells per travelling distance decreases in more complex test scenarios. 

\begin{figure*}[t]
\centering
\begin{subfigure}[b]{0.327\textwidth}
        \centering
        \includegraphics[keepaspectratio,width=4.5cm]{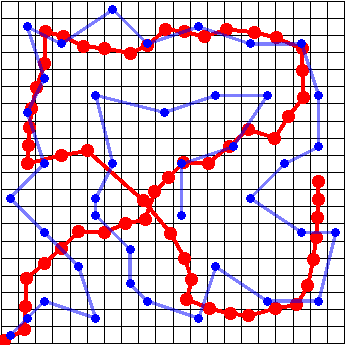}
        \caption{}
         \label{CompMap1}
\end{subfigure}
\begin{subfigure}[b]{0.327\textwidth}
        \centering
        \includegraphics[keepaspectratio,width=4.5cm]{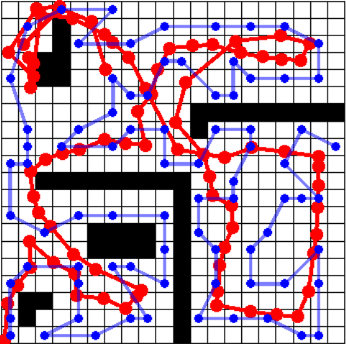}
         \caption{}
         \label{CompMap2}
\end{subfigure}
\begin{subfigure}[b]{0.327\textwidth}
        \centering
        \includegraphics[keepaspectratio,width=4.5cm]{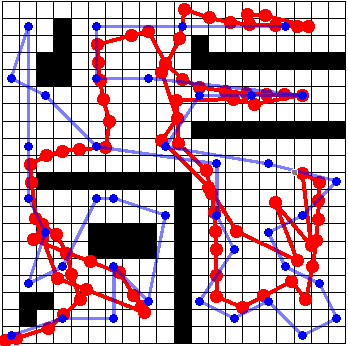}
         \caption{}
         \label{CompMap3}
\end{subfigure}
\begin{subfigure}[b]{0.327\textwidth}
        \centering
        \includegraphics[keepaspectratio,width=4.5cm]{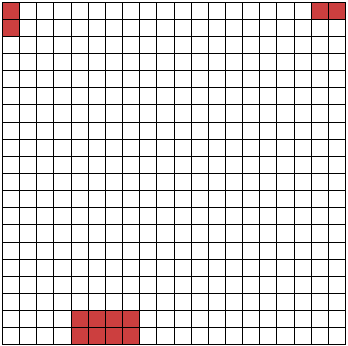}
         \caption{}
         \label{MapCoverage1}
\end{subfigure}
\begin{subfigure}[b]{0.327\textwidth}
        \centering
        \includegraphics[keepaspectratio,width=4.5cm]{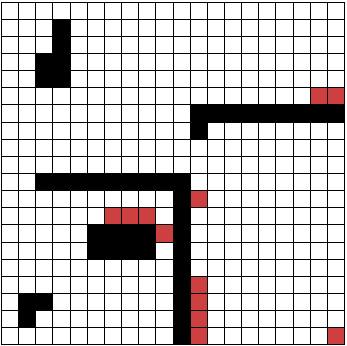}
         \caption{}
         \label{MapCoverage2}
\end{subfigure}
\begin{subfigure}[b]{0.327\textwidth}
        \centering
        \includegraphics[keepaspectratio,width=4.5cm]{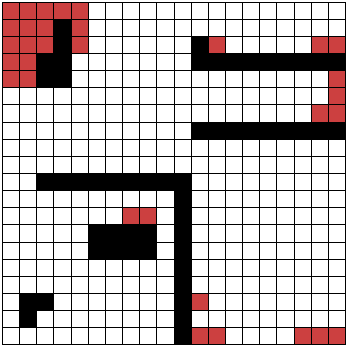}
         \caption{}
         \label{MapCoverage3}
\end{subfigure}
\caption{ (a)-(c) A sample of the generated path by the proposed HIPP (the res curves) as compared to the benchmark solution (the blue curves), in three test scenarios. These paths are those which are generated the best results (as the maximum equivalent number of identified cells per unit of travelling distance) out of the ten runs of test scenarios. (d)-(f) the cells (in red) that HIPP-generated paths are not confidently scanned as empty cells.}
% \caption{Total cells covered by the most confident map by the LiDar during the mapping of different test scenarios }
\label{fig:simulationResults1}
\end{figure*}

Fig. \ref{fig:simulationResults1} compares the best-generated paths by the proposed HIPP algorithm against the best result by solving the posterior problem out of ten runs of the algorithms. The resulting maps of HIPP are also illustrated in Fig. \ref{MapCoverage1}-\ref{MapCoverage3} indicating the unmapped parts as red cells. The readers are also referred to the prepared video on \url{https://youtu.be/wnBgTZxwG_o} where the performance of HIPP for all three test scenarios is demonstrated. 

\begin{figure*}[!]
\centering
\begin{subfigure}[b]{0.327\textwidth}
        \centering
        \includegraphics[keepaspectratio,width=4.5cm]{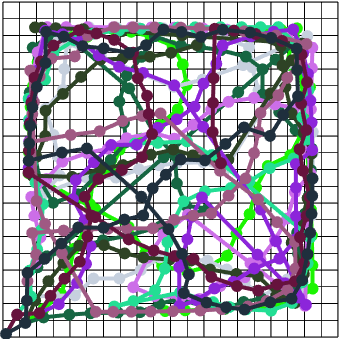}
%         \caption{Map 1. HIPP}
         \caption{}
         \label{Map1HIPP}
\end{subfigure}
\begin{subfigure}[b]{0.327\textwidth}
        \centering
        \includegraphics[keepaspectratio,width=4.5cm]{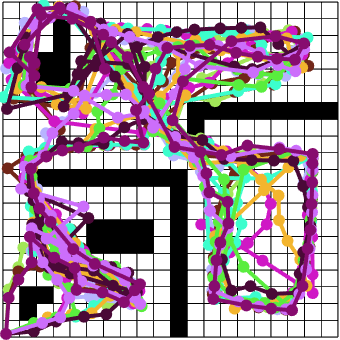}
         \caption{}
%         \caption{Map 2. Posterior}
         \label{Map2HIPP}
\end{subfigure}
\begin{subfigure}[b]{0.327\textwidth}
        \centering
        \includegraphics[keepaspectratio,width=4.5cm]{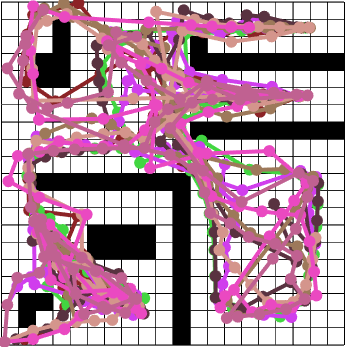}
         \caption{}
%         \caption{Map 3. HIPP}
         \label{Map3HIPP}
\end{subfigure}
\caption{Variation of generated paths by ten runs of the proposed HIPP algorithm for the three test scenarios. These results show the similarities between different runs of each test scenario in terms of the travelling paths of the robot. }
\label{fig:simulationResults2}
\end{figure*}

Fig. \ref{fig:simulationResults2} visualizes the variation of the generated paths by HIPP over ten different runs which shows the consistency of the algorithm. For example, it shows that HIPP ignores exploring the central areas of rooms and the areas close to the walls.

\par 
\begin{table}[t]
\caption{The statistics of total travelling distance (TTD), number of identified cells and that per each unit of travelling distance (TD) for ten times running of Algorithms \ref{alg:cap1} and \ref{alg:cap2} for test scenarios 1 to 3. An equivalent number of identified cells, calculated by placing the ideal sensor of Fig.\ref{OGM_idealSensor} at the generated way-points by HIPP, provides a realistic comparison, as explained in section \ref{results}.}
\label{tab5}
\centering
\begin{tabular}{p{0.02\linewidth}p{0.13\linewidth}p{0.05\linewidth}p{0.08\linewidth}p{0.10\linewidth}p{0.11\linewidth}p{0.12\linewidth}}
\hline
& & TTD (m) & No. of identified cells & No. of identified cells/TD (cells/m) & Equivalent no. of identified cells & Equivalent no. identified cells/TD (cells/m)\\
\hline

\multirow{10}{*}{\rotatebox[origin=c]{90}{\textbf{\textcolor{red}{Test Scenario 1}}}}
    &\multicolumn{6}{c}{\cellcolor{gray!20} Algorithm \ref{alg:cap1} (HIPP) results}\\  
    & Average & 48.9 & 384.5 & {8.1} & 233.5 & \textbf{4.9} \\
    & Std. Deviation & 12.21 & 13.6  & 2.4  & 15.8 & 1.1\\
    & Median  &  49.3 & 388.0 & 7.7 & 231.0 & 4.7 \\
    \cline{2-7}
    
    &\multicolumn{6}{c}{\cellcolor{gray!20} Algorithm \ref{alg:cap2} (Posterior) results, i.e., a benchmark solution}\\
    & Average & 58.1 & 362.0 & \textbf{6.2} & - & - \\
    & Std. Deviation & 1.0 & 6.4  & 0.1  & - & -\\
    & Median  &  58.4 & 364.0 & 6.2 & - & - \\
\hline
\multirow{10}{*}{\rotatebox[origin=c]{90}{\textbf{\textcolor{red}{Test Scenario 2}}}}
    &\multicolumn{6}{c}{\cellcolor{gray!20} Algorithm \ref{alg:cap1} (HIPP) results}\\
    & Average & 78.8 & 340.0 & {4.4} & 262.0 & \textbf{3.4}\\
    & Std. Deviation & 8.8 & 7.4  & 0.6  & 10.0 & 0.4\\
    & Median  &  80.6 & 343.0 & 4.3 & 262.0 & 3.3 \\
    \cline{2-7}

    &\multicolumn{6}{c}{\cellcolor{gray!20} Algorithm \ref{alg:cap2} (Posterior) results, i.e., a benchmark solution}\\
    & Average & 83.8& 353.0 & \textbf{4.2} & - & - \\
    & Std. Deviation & 2.9 & 1.7  & 0.1  & - & -\\
    & Median  &  84.4 & 354.0 & 4.2 & - & - \\
\hline
\multirow{10}{*}{\rotatebox[origin=c]{90}{\textbf{\textcolor{red}{Test Scenario 3}}}}
    &\multicolumn{6}{c}{\cellcolor{gray!20} Algorithm \ref{alg:cap1} (HIPP) results}\\
    &Average & 80.7 & 305.5 & {3.8} & 241.0 & \textbf{3.0} \\
    &Std. Deviation & 7.6 & 27.2  & 0.6  & 18.7 & 0.4\\
    &Median  &  80.0 & 313.5 & 3.9 & 244.0 & 3.1 \\
    \cline{2-7}

    &\multicolumn{6}{c}{\cellcolor{gray!20} Algorithm \ref{alg:cap2} (Posterior) results, i.e., a benchmark solution}\\    &Average & 72.5 & 311.6 & \textbf{4.3}& - & - \\
    &Std. Deviation & 2.8 & 8.7  & 0.2  & - & -\\
    &Median  &  72.7 & 310.0 & 4.3 & - & - \\
\hline
\end{tabular}
\end{table}

\begin{figure}[t]
   \begin{subfigure}[b]{0.45\textwidth}
   \centering
   \includegraphics[keepaspectratio,width=7.0cm]{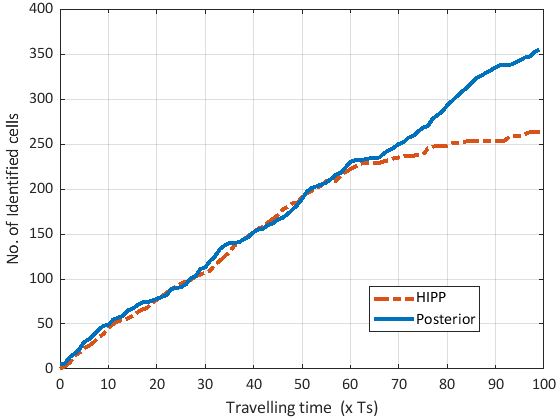}
   
   \vspace{-5pt}
   \caption{}
   \label{comparisionsNoCells}
   \end{subfigure}

   \begin{subfigure}[b]{0.45\textwidth}
   \centering
   \includegraphics[keepaspectratio,width=7.0cm]{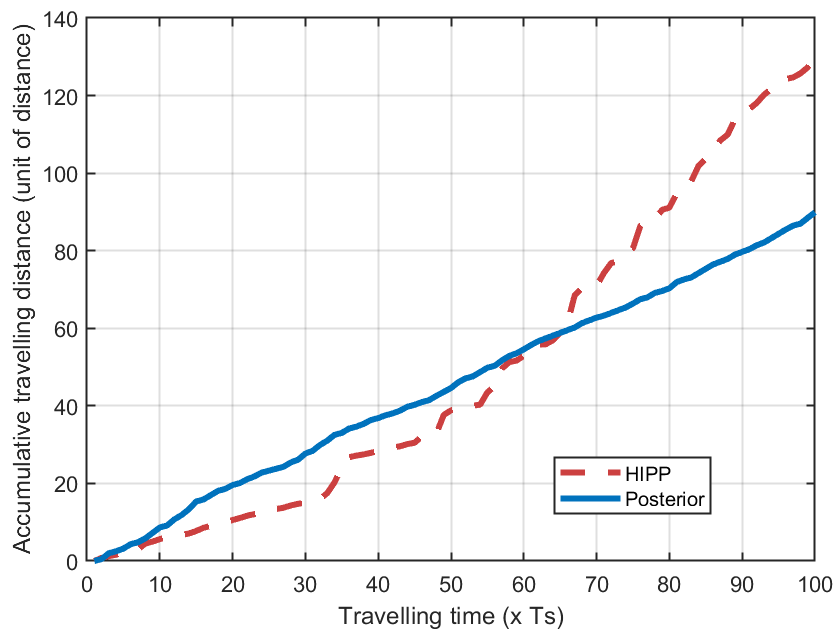}
   
   \vspace{-5pt}
   \caption{}
   \label{comparisionsTTD}
   \end{subfigure}

   \begin{subfigure}[b]{0.45\textwidth}
   \centering
   \includegraphics[keepaspectratio,width=7.0cm]{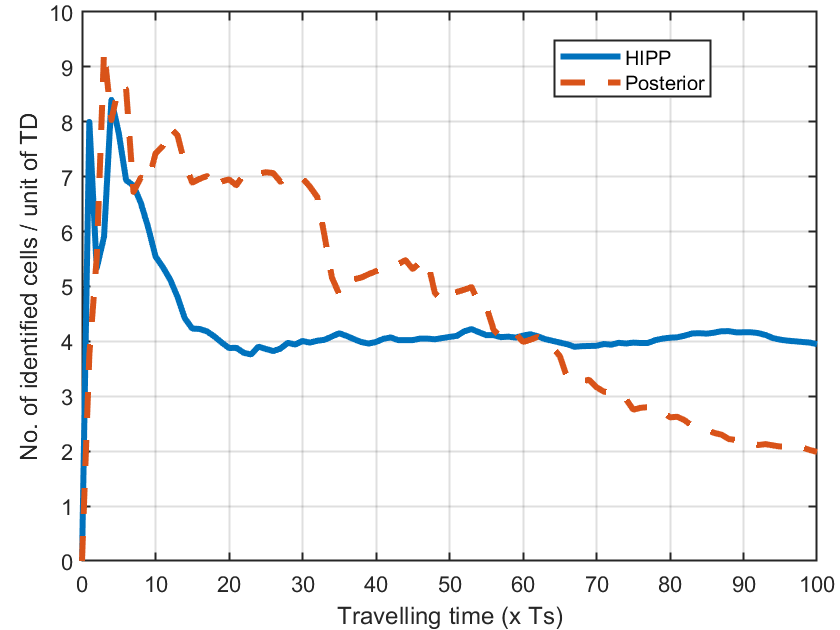}
   
   \vspace{-5pt}    
   \caption{}
  \label{comparisionsPerTD}
   \end{subfigure}

 \caption{Comparison of the (a) accumulative number of identified cells up to each travelling time (in terms of the number of sampling times $T_s$), (b) accumulative travelling distance and (c) the instantaneously identified cells per variation of travelling distance for the Test Scenario 2.}
 \label{fig:comparisions}
\end{figure}

The instantaneous behaviour of the proposed HIPP algorithm over the travelling time is compared in Fig. \ref{fig:comparisions} to such results of the benchmark solution. These results are for test scenario 2, however, the other scenarios follow a similar trend. 

As shown in Fig.\ref{comparisionsNoCells}, the accumulative number of identified cells by HIPP follows an almost linear increase by travelling time before it starts being saturated at around $60$ sampling times $T_s$. In other words, HIPP follows the performance of the benchmark solution over a wide range of operating time before being outperformed. 

However, it is shown in Fig.\ref{comparisionsTTD} that within the same period of $60$ sampling times, the robot travels less by HIPP than the benchmark solution. This means that over the first half of the travelling time, HIPP outperforms the benchmark solution in terms of the normalised identified cells by travelling distance. Fig. \ref{comparisionsPerTD} illustrates this comparison measure clearer where the number of identified cells per unit of travelling distance is plotted against the travelling time. This means that HIPP implicitly prioritises the wider areas where the mapping gains are higher than narrow passages.

In addition, it is observed in Fig. \ref{comparisionsPerTD} that there is a sharp rise in the number of identified cells per travelling distance by the solution of the posterior problem before reaching a saturated state after around $15$ seconds. This is because of scanning multiple cells with a small movement. The saturated $4$ cells per unit of travelling distance match the range of the sensor which is $1.5$ cells or equivalently $75\,$cm which is an indicator of finding a near-to-global optimal solution. 

The normalised number of identified cells by HIPP, on the other hand, continuously decreases over travelling time before reaching $2$ cells per unit of travelling distance at the end of the test scenario. 

\section{Conclusion}
This paper presents a novel heuristic algorithm (which is called HIPP) for planning the exploration path of a single robot to efficiently map an unknown non-convex environment and make a map out of it. The robot has a LiDAR sensor and is equipped with the SLAM algorithm, however, it does not have knowledge about the map or the final destination. The efficiency of the algorithm is measured by the number of identified cells per unit of travelling distance. 

A posterior problem (i.e., where the robot remaps a known map) is formulated and its solution is considered as the benchmark to evaluate the performance of HIPP. It is shown that such a problem is a complex discontinuous problem. This paper also proposes a novel solver for the posterior problem to find a near-optimal solution to remapping a known non-convex area where the final destination is not given.

The simulation results show that the formulated IPP problem has potentially multiple close sub-optimal solutions which are probably close to the globally optimal path.

We could not find a direct solution to the posterior problem, however, we could develop a heuristic solver to this problem. The results show that this solver finds statistically similar results with a narrow distribution. However, it is observed that the posterior problem has potentially multiple close optimal solutions.

Overall, the simulation results for different test scenarios show that the proposed HIPP achieves $70-80\%$ of the benchmark solution in terms of the number of identified cells per travelling distance. Moreover, whilst the target map is expanded linearly and with a similar rate as the benchmark solution over the first $60\%$ of the exploration time, the robot travels less using HIPP for this first period of exploration. This behaviour is desired for such information-gathering algorithms indicating that the proposed HIPP creates a significant portion of the map in the shortest possible time (or travelling distance) by an implicit prioritisation which is of significant importance in hazardous applications 

%\bibliographystyle{IEEEtranN}
%\bibliography{Bibliography.bib}

\printbibliography %Prints bibliography
\end{document}